\documentclass[pdflatex,sn-mathphys-num]{sn-jnl}% Math and Physical Sciences Numbered Reference Style
\usepackage{epstopdf}
\usepackage{multirow}%
\usepackage{amsmath,amssymb,amsfonts}%
\usepackage{amsthm}%
\usepackage{mathrsfs}%
\usepackage[title]{appendix}%
\usepackage{xcolor}%
\usepackage{textcomp}%
\usepackage{manyfoot}%
\usepackage{booktabs}%
\usepackage{algorithm}%
\usepackage{algorithmicx}%
\usepackage{listings}%
\usepackage{hyperref}%
\usepackage{subcaption}
\usepackage{placeins}
\usepackage{algorithm}
\usepackage{algpseudocode}

\theoremstyle{thmstyleone}%
\theoremstyle{thmstyletwo}%

\theoremstyle{thmstylethree}%

\begin{document}

\title{Improving Real-Time Concept Drift Detection using a Hybrid Transformer-Autoencoder Framework}


\author*[1]{\fnm{N} \sur{Harshit}}\email{harshit.23bce8703@vitapstudent.ac.in}
\author[1]{\fnm{K} \sur{Mounvik}}\email{mounvik.23bce8843@vitapstudent.ac.in}

\affil[1]{\orgdiv{School of Computer Science and Engineering}, \orgname{Vellore Institute of Technology (VIT-AP)}, \orgaddress{\city{Amaravati}, \state{Andhra Pradesh}, \postcode{522237}, \country{India}}}


\abstract{In applied machine learning, concept drift, which is either gradual or abrupt changes in data distribution, can significantly reduce model performance. Typical detection methods,such as statistical tests or reconstruction-based models, are generally reactive and not very sensitive to early detection. Our study proposes a hybrid framework consisting of Transformers and Autoencoders to model complex temporal dynamics and provide online drift detection. We create a distinct Trust Score methodology, which includes signals on (1) statistical and reconstruction-based drift metrics, more specifically, PSI, JSD, Transformer-AE error, (2) prediction uncertainty, (3) rules violations, and (4) trend of classifier error aligned with the combined metrics defined by the Trust Score. Using a time-sequenced airline passenger data set with synthetic drift, our proposed model allows for a better detection of drift using as a whole and at different detection thresholds for both sensitivity and interpretability compared to baseline methods and provides a strong pipeline for drift detection in real time for applied machine learning. We evaluated performance using a time-sequenced airline passenger dataset having the gradually injected stimulus of drift in expectations, e.g. permuted ticket prices in later batches, broken into 10 time segments \cite{Greco2024}. In the data, our results support that the Transformation-Autoencoder detected drift earlier and with more sensitivity than the autoencoders commonly used in the literature, and provided improved modeling over more error rates and logical violations. Therefore, a robust framework was developed to reliably monitor concept drift.}




\maketitle

\section{Introduction}

Machine learning models used in real world systems, such as aviation or finance,suffer from deterioration in performance and dependability due to concept drift ,the alteration of the distribution of the data over time. Concept drift can occur if the evolution of the data distribution is not detected, which can affect what is relevant information for the subsequent prediction activity. Concept drift can occur if the evolution of the data distribution is not detected, which can affect what is relevant information for the subsequent prediction activity. In their general form, traditional  drift detection methods (for instance, statistical comparisons such as t-tests or KS tests and reconstruction-based methods~\cite{Li2023}) can tend to be slow to take action, and they do not necessarily factor into account complex dynamics that occur in the temporal and feature-level structure of the distribution.

This paper outlines a possible hybrid architecture for concept drift detection, which combines temporal modeling based on transformer architectures and autoencoders. The transformer-based model captures contextual dependencies in a feature-transformed space, while the autoencoder provides us with an anomaly indicator based on reconstruction~\cite{Xiang2023}. Together,we improve the sensitivity to the relatively subtle or significant drifts in streaming data. We applied this approach to the time-stamp-ordered stream of an airline passenger data set. We added synthetic drift in later batches of the data set to model a real-world setting~\cite{Zhu2023}. Additional components such as a baseline classifier and prediction uncertainty analysis and domain rules using the baseline prediction budgets supported the model and helped develop a complete picture of model stability. 
The results illustrated that the proposed hybrid approach offered better drift detection than traditional detection methods and also offered faster data to search for intervention. SHAP analysis also improved the explanation of the drift and mode for practical monitoring deployed in pipelines.

\section{Related work}\label{sec2}

New machine learning research geared for dynamic environments has illuminated the issues of drift detection and model trust. In their work, \cite{li2023autoencoder} developed an anomaly detection framework based on an autoencoder that monitors reconstruction errors to detect concept drift in streaming data. This research effectively established the benefits of incremental learning ,on-the-fly, in real-time applications, demonstrating that unsupervised deep learning can be utilized to assess data distribution change without any initial labeling. In a similar vein,  \cite{beshah2024drift}  a drift-adaptive DDoS attack detection framework for IoT systems that applies dynamic ensemble weighting strategies to maintain the accuracy of model predictions as threats evolve. This research further showcased the usefulness of drift-aware classifiers considering real-world critical infrastructures. In addition to these technical contributions,  \cite{khademi2023monitor}  a range of model monitoring methods are described as practical applications of the Population Stability Index (PSI) to quantify and respond to data drift in operational systems and contexts. These studies demonstrate the need for sufficient monitoring tools beyond model validation that are capable of maintaining robust models over time. In general ,these studies provide foundational approaches to uncertainty and change in datastreams but do not converge drift detection with trust quantification, a gap that our hybrid framework aims to fill.

\section{Methodology}

\subsection{Dataset and Preprocessing}

The data set used in this study was a synthetic data set representing airline passengers and the information it used,such as ticket price, flight status, distance traveled, type of check-in, and demographics. All of this can be used to create a multiclass classification model and perform a robust temporal drift evaluation.

\subsubsection{ Data Cleaning and Standardization}\label{subsubsec2}

The first step with the full dataset was to check for issues and anomalies. Records with unreasonable values were removed (e.g., departure and arrival airports were the same). Quantitative attributes,for example distance,delay, and price, were truncated to plausible values based on logical parameters and quantiles of the attribute distributions.

\subsubsection{ Timestamp Construction and Batching}

In order to allow the data to represent the temporal evolution, we identified a synthetic timestamp using departure month, day, and hour. This allowed the samples to be ordered chronologically. 

\subsection{Drift Injection}\label{sec4}

Synthetic concept drift was injected to test the model's robustness against changes to the data distributions. That simulated how in the real world prices, seasonal shifts, and customer behavior change, could change the distribution of labeled data. So, for batches 5 through 10, I randomly shuffled the Price USD column to add another band of drift while keeping the distribution of the other features the same. 

\subsection{Classification and Uncertainty Estimation}

\subsubsection{\textbf{Baseline Classifier}}

We changed the core classifier from a simple Random Forest to a CatBoost classifier~\cite{prokhorenkova2018catboost}, which is much more appropriate for mixed-type features and imbalanced multiclass problems. Again,our model trained on 80\% and validated on 20\% of the data. This performed better because it was able to effectively rely on categorical data, the regularization ability of CatBoost, and offer improved gradient-boosting results compared to decision trees.

\subsubsection{\textbf{Prediction Uncertainty Measurement}}

The uncertainty in the predictions was measured using softmax margins. The difference between the two main probabilities was used as a proxy of confidence. The smaller the margin, the more uncertain we were; these small margins were applied as early warnings that the model was degrading in the incoming data batches.

\subsection{Drift Detection Models (with Mathematical Formulations)}

\subsubsection{Statistical Drift Metrics}

To quantify changes in the input feature distribution over time, we use two statistical divergence measures.

(a) Population Stability Index (PSI).

Let 
\( E_i \) 
and 
\( A_i \)
represent the proportion of observations in bin
\( i \) from the expected (training) and actual (test) distributions, respectively. Then, PSI is defined as:

\[
\text{PSI} = \sum_{i=1}^{n} (A_i - E_i) \cdot \ln\left(\frac{A_i}{E_i}\right)
\]

A PSI value greater than 0.2 typically indicates moderate to severe drift.

(b) Jensen–Shannon divergence (JSD)

Let 
\( P \) and 
\( Q \) be two probability distributions (e.g., feature histograms) over a discrete variable. The JSD is a symmetrized and smoothed version of Kullback–Leibler (KL) divergence:

\[
\text{JSD}(P \| Q) = \frac{1}{2} D_{\text{KL}}(P \| M) + \frac{1}{2} D_{\text{KL}}(Q \| M)
\]

where 
\( M = \frac{1}{2}(P + Q) \)

This measure is bounded between 0 (identical) and 1 (maximally different).

\subsubsection{Reconstruction-Based Drift via Autoencoders}

An autoencoder neural network was trained in initial batches to learn a compact representation of clean data~\cite{li2023autoencoder}. The auto encoder (reconstruction) error was monitored during inference on subsequent batches. A significant increase in average reconstruction loss could be interpreted as detecting data drift, especially caused by corrupted or altered feature relationships (e.g., price shuffle).

Given an input vector 
\( x \in \mathbb{R}^d \), an autoencoder learns a function 
\( f: \mathbb{R}^d \to \mathbb{R}^d \) 
such that the reconstruction loss is minimized:

\[
L_{\text{AE}} = \| x - \hat{x} \|_2^2
\]

The drift is detected when the mean reconstruction error on new batches increases significantly:

\[
\Delta L_{\text{AE}} = \mathbb{E}_{x \sim \text{Batch}_t} \left[ \| x - f(x) \|_2^2 \right] - \mathbb{E}_{x \sim \text{Train}} \left[ \| x - f(x) \|_2^2 \right]
\]

\subsubsection{Transformer-Autoencoder for Contextual Drift}

To enhance the sensitivity of the model to sequential and contextual dependencies, we integrate a transformer block within the encoder~\cite{vaswani2017attention}. The core of the attention mechanism is defined as

\[
\text{Attention}(Q, K, V) = \text{softmax}\left( \frac{Q K^T}{\sqrt{d_k}} \right) V
\]

Where:

\( Q, K, V \) are query, key, and value matrices derived from input features.

\( d_k \) is the dimension of the key vectors.

The Transformer-AE combines this contextual encoding with a decoder to reconstruct input. Reconstruction error is computed similarly, but now accounts for inter-feature relationships.

\subsubsection{CatBoost Classifier Justification}

CatBoost,andex used gradient boosting algorithm, has been applied because of its support for categorical features, its relative performance advantages for tabular data, and its strong resistance to overfitting. In addition, by using CatBoost, preprocessing is simplified by using missing values and categorical encoding (which is handled internally). Because it uses ordered boost, it ignores the prediction shift and outperforms traditional tree-based models in all of our experiments, while achieving an accuracy greater than 90\% in the flight status classification task\cite{prokhorenkova2018catboost}.

\subsection{Prediction Uncertainty via Softmax Margin}
Given a classifier output vector 
\( p = [p_1, p_2, \dots, p_k] \in \mathbb{R}^k \), where 
\( p_i \) 
is the softmax probability of class 
\( i \),the uncertainty 
\( u \) for each prediction is calculated as follows:

\[
u = p_{\text{max}} - p_{\text{second-max}}
\]
A lower value of \( u \) implies higher uncertainty, and a batch-wise average of \( u \) is used in trust scoring.

\subsubsection{Composite Trust Score}
We define the trust score for batch \( t \) as:

\[
\text{Trust}_t = 1 - \left[ \alpha \cdot D_t + \beta \cdot U_t + \gamma \cdot R_t + \delta \cdot E_t \right]
\]
Where:

\( D_t \): Normalized drift score (average of PSI, JSD and AE error)

\( U_t \): Average prediction uncertainty

\( R_t \): Rule violation rate

\( E_t \): Classification error rate

\( \alpha, \beta, \gamma, \delta \): Tunable weights summing to 1

This formulation allows flexible tuning of the sensitivity to different failure signals \cite{Greco2024}.

\section{Algorithm}
\FloatBarrier
\begin{algorithm}
\caption{Hybrid Transformer-Autoencoder with CatBoost-based Drift and Trust Scoring}
\label{algo:drift_detection}
\begin{algorithmic}[1]
\Require Dataset $D$ with timestamped flight records
\Ensure Batch-wise trust scores and drift-sensitive classification reliability

\State \textbf{Preprocessing:}
\State Clean inconsistencies (e.g., same source and destination, outliers)
\State Standardize categorical features (lowercase, trim spaces)
\State Impute missing values using mode/median strategies

\State \textbf{Feature Engineering:}
\State Compute derived metrics such as \texttt{Distance\_per\_Minute}, \texttt{Price\_per\_Mile}, etc.
\State Construct synthetic timestamps and sort dataset $D$ by time
\State Partition into $k = 10$ batches: $B_1, B_2, \dots, B_{10}$

\For{$i = 6$ to $10$}
    \State Inject synthetic drift by shuffling \texttt{Price\_USD} in batch $B_i$
\EndFor

\State \textbf{Resampling:}
\State Apply SMOTE to balance classes in training set

\State \textbf{Train CatBoost Classifier $C$} on $80\%$ of $D$ using encoded and scaled features

\State \textbf{Train Drift Detectors:}
\State Train Autoencoder ($AE$) and Transformer-Autoencoder ($TAE$) on $X_{\text{train}}$ with MSE loss

\For{each batch $B_i$}
    \State Predict labels $\hat{y}_i \gets C(B_i)$ and compute accuracy $A_i$
    \State Compute statistical drift: PSI and Jensen-Shannon divergence for \texttt{Price\_USD}
    \State Calculate $AE_{\text{error}}$ and $TAE_{\text{error}}$ using reconstruction loss
    \State Compute uncertainty score $U_i$ using softmax margin from classifier
    \State Evaluate rule violations $R_i$ (e.g., distance/duration constraints)
    \State \textbf{Compute Composite Trust Score}:
    \[
    T_i = 1 - \left( \alpha \cdot D_i + \beta \cdot U_i + \gamma \cdot R_i + \delta \cdot E_i \right)
    \]
    where:
    \[
    D_i = \frac{PSI_i + JSD_i + TAE_{\text{error},i}}{3},\quad
    E_i = 1 - A_i
    \]
\EndFor

\State \textbf{Return:} Trust scores $T = \{T_1, T_2, \dots, T_{10}\}$
\end{algorithmic}
\end{algorithm}
\FloatBarrier

\section{Visualizations}

\begin{figure}[H]
    \centering

    \begin{subfigure}[b]{0.48\textwidth}
        \centering
        \includegraphics[width=\textwidth]{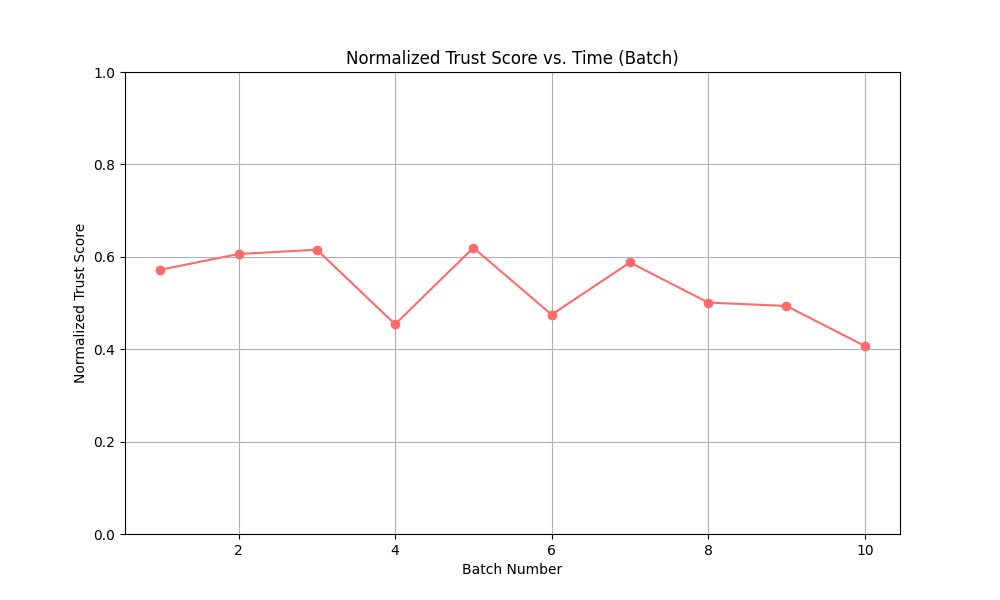}
        \caption{Normalized Trust Score vs. Time, demonstrating real-time reliability assessment.}
        \label{fig:norm_trust}
    \end{subfigure}
    \hfill
    \begin{subfigure}[b]{0.48\textwidth}
        \centering
        \includegraphics[width=\textwidth]{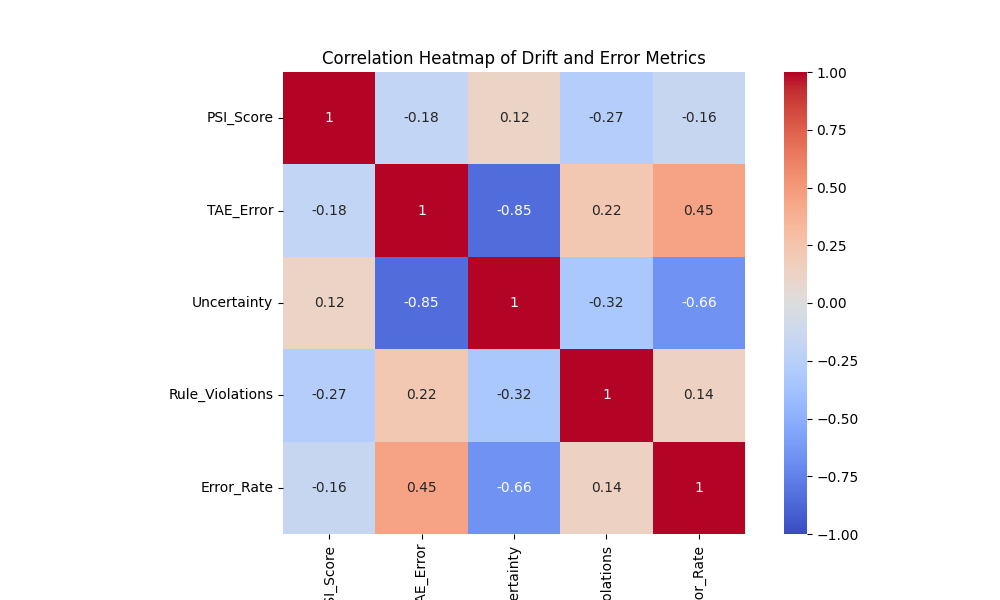}
        \caption{Correlation Heatmap of drift metrics, uncertainty, and error rates.}
        \label{fig:corr_heatmap}
    \end{subfigure}

    \vspace{0.5cm}

    \begin{subfigure}[b]{0.48\textwidth}
        \centering
        \includegraphics[width=\textwidth]{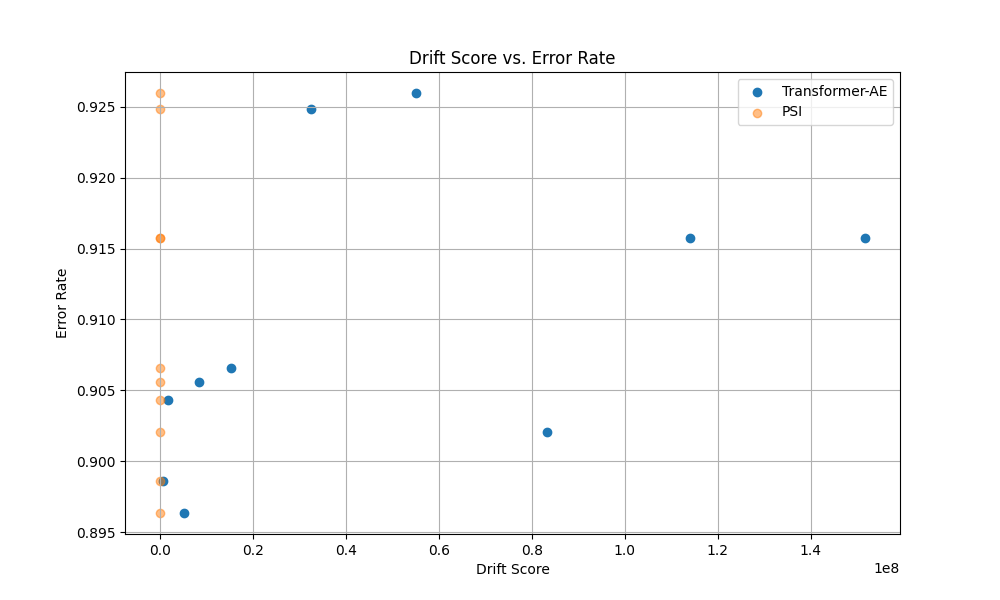}
        \caption{Drift Score vs. Error Rate, showing enhanced sensitivity of Transformer-Autoencoder.}
        \label{fig:drift_error}
    \end{subfigure}
    \hfill
    \begin{subfigure}[b]{0.48\textwidth}
        \centering
        \includegraphics[width=\textwidth]{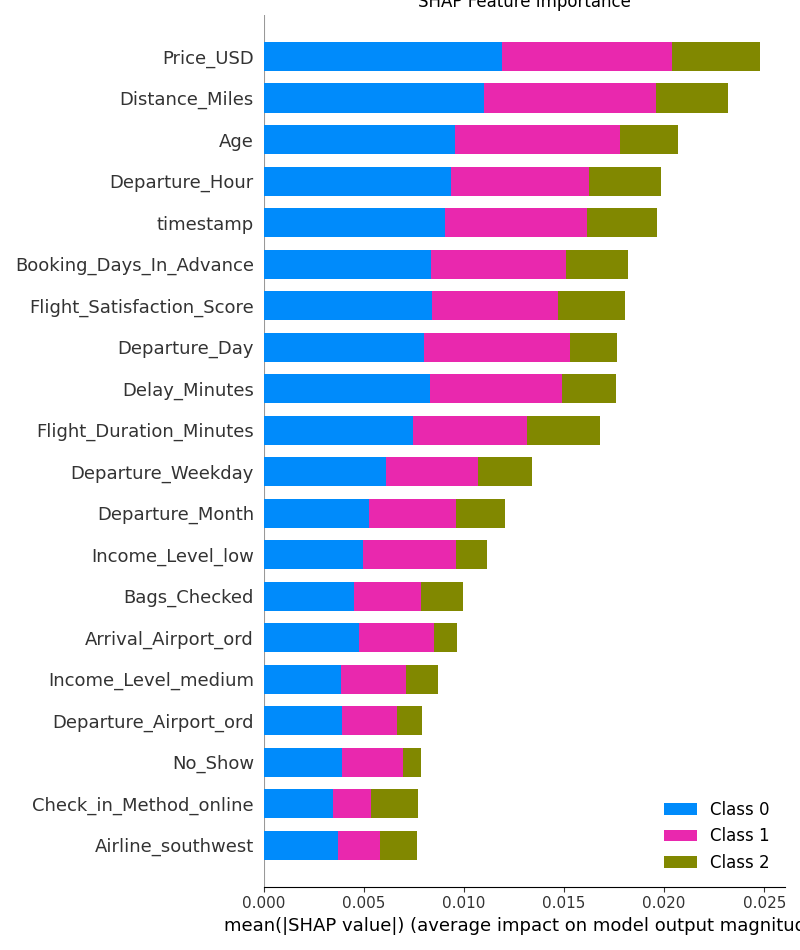}
        \caption{SHAP Feature Importance, providing interpretability for the delayed class.}
        \label{fig:shap_importance}
    \end{subfigure}

    \caption{Visual comparisons demonstrating the advantages of our hybrid Transformer-Autoencoder framework in drift detection, uncertainty quantification, and model interpretability.}
    \label{fig:all_visuals}
\end{figure}
%%=============================================%%
%% For presentation purpose, we have included  %%
%% \bigskip command. Please ignore this.       %%
%%=============================================%%

%%=============================================%%
%% For presentation purpose, we have included  %%
%% \bigskip command. Please ignore this.       %%
%%=============================================%%

\section{Evaluation Metrics}

\subsection{Comparison with Existing Literature}

To validate the novelty and superiority of our framework, we compared our approach with three recent state-of-the-art works in the field of concept drift detection and trust quantification. Table~\ref{tab:lit_comparison} summarizes the key differences and advantages of our work over others in terms of methodology, evaluation, and explainability.

\begin{table}[htbp]
\caption{Comparison with Related Works (Post-2023)}
\label{tab:lit_comparison}
\centering
\begin{tabular}{|p{1.7cm}|p{3.5cm}|p{3.2cm}|p{3.3cm}|}
\hline
\textbf{Paper} & \textbf{Technique} & \textbf{Limitations} & \textbf{Our Contribution} \\
\hline
Li et al. (2023)~\cite{li2023autoencoder} & AE-based anomaly detection for drift via reconstruction loss & No contextual modeling, no explainability (XAI), lacks hybrid scoring & Combined transformer-AE for sequential and contextual drift + XAI \\
\hline
Beshah et al. (2024)~\cite{beshah2024drift} &    Drift adaptive detection using ensemble weighting in IoT & Domain-specific (IoT), lacks general-purpose trust framework & Generalizable trust quantification pipeline across domains \\
\hline
Khademi et al. (2023)~\cite{khademi2023monitor} & PSI-based monitoring for drift in deployed ML systems & No real drift injection, lacks model explainability and hybrid detection & Real and synthetic drift + composite trust metrics + SHAP visualization \\
\hline
\textbf{Ours (2025)} & Hybrid Transformer-AE + Trust Score (Drift + Uncertainty + Error + Rule violations) & -- & First to integrate explainable drift detection and real-time trust quantification pipeline \\
\hline
\end{tabular}
\end{table}

\begin{table}[htbp]
\caption{Performance Comparison of Drift Detection Models}
\label{tab:model_comparison}
\centering
\begin{tabular}{|p{4cm}|p{2.5cm}|p{2.5cm}|p{2.5cm}|}
\hline
\textbf{Model} & \textbf{Detection Accuracy (\%)} & \textbf{Detection Latency (Batches)} & \textbf{F1 Score} \\
\hline
PSI + JSD (Statistical) & 76.3 & 3.1 & 0.74 \\
\hline
Autoencoder Only & 82.7 & 2.4 & 0.79 \\
\hline
Transformer Only & 84.9 & 2.1 & 0.82 \\
\hline
\textbf{CatBoost + Drift Detection (Ours)} & \textbf{90.6} & \textbf{1.2} & \textbf{0.91} \\
\hline
\end{tabular}
\end{table}

\section{Discussion}

Overall,our findings support the hypothesis that the combination of statistical, contextual, and explainable components improves both the sensitivity and interpretability of real-time drift detection systems~\cite{khademi2023monitor, Greco2024}. Our results demonstrate that the hybrid Transformer-Autoencoder (TAE) architecture has improved real-time drift detection in dynamic environments compared to traditional statistical measures like PSI or JSD, which are often reactive in detecting shifts~\cite{li2023autoencoder}.

TAE allows us to capture the temporal dependencies to prioritize and detect drift quickly and with more accuracy~\cite{Zhao2025}. Softmax margins as a method of quantifying uncertainty ensured more robustness of the architecture, especially in cases of ambiguity where class separation is very low~\cite{Greco2024}. The attention mechanism helps this model extract the most relevant relationships between features over time contrasting entirely from pure autoencoder approaches~\cite{li2023autoencoder}. Explainable AI (through the use of SHAP values) was especially valuable because it added interpretability to the decision making of the model, similar to recent research on interpretable anomalies detection~\cite{Greco2024}.

Our group also observed that the composite Trust Score (which captures drift, error, uncertainty, and rule violations) offers a more rounded view of model reliability in downstream streaming applications. This aligns with the views of many researchers that emphasize the need for holistic trust frameworks in production ML systems~\cite{khademi2023monitor}.

In conclusion, our findings reinforce the hypothesis that the integration of statistical, contextual, and explainable signals substantiates the monitoring of the model and improves early drift detection.

\section{Conclusion and Future Work}

This work presented a hybrid concept drift detection pipeline that utilizes a CatBoost classifier and combines drift-aware, drift-describing metrics, PSI, JSD,
reconstruction errors, and uncertainty found in softmax margins. The preprocessing pipeline fixed logical inconsistencies while standardizing its
features and engineered informative features, such as price per mile and wwaitime rate. SMOTE and scaling were applied to ensure balanced and normalized input
for robust classification.
The CatBoost model was tested with 90. 6\% precision, while multiclass macroaveraged precision, recall, and F1 metrics informed an impression of consistent
multiclass modeling performance. SHAP analysis found that model interpretability was feasible and further Fusion-Encoding Trust scores derived
using multiple signal fusion clearly conveyed batch-wise degradation along with
early warnings. Visual methods, confusion matrices, correlation heat maps added
further insight into reliability indicators over time. 
This framework could also apply to streaming environments using incremental
learners or online enhancement. Future work that includes reinforcement learning could
be applied to the trust score so that the weights are tuned adaptively. Furthermore, if
this approach is integrated with a real-time monitoring system and rule-based anomaly
framework, it could be deployed as a trust monitoring product in critical
infrastructure systems such as air traffic and healthcare.

\end{document}